# Optimal path for Biomedical Text Summarization Using Pointer-GPT

Hyunkyung Han[1*]  Jaesik Choi[12]

[1] Korea Advanced Institute of Science and Technology  [2] INEEJI

## Abstract

Biomedical text summarization is a critical tool that enables clinicians to effectively ascertain patient status. Traditionally, text summarization has been accomplished with transformer models, which are capable of compressing long documents into brief summaries. However, transformer models are known to be among the most challenging natural language processing (NLP) tasks. Specifically, GPT models have a tendency to generate factual errors, lack context, and oversimplify words. To address these limitations, we replaced the attention mechanism in the GPT model with a pointer network. This modification was designed to preserve the core values of the original text during the summarization process. The effectiveness of the Pointer-GPT model was evaluated using the ROUGE score. The results demonstrated that Pointer-GPT outperformed the original GPT model. These findings suggest that pointer networks can be a valuable addition to EMR systems and can provide clinicians with more accurate and informative summaries of patient medical records. This research has the potential to usher in a new paradigm in EMR systems and to revolutionize the way that clinicians interact with patient medical records.

## 1. Introduction

Biomedical text summarization is critical for clinicians to recognize patient status effectively. [1] And summarization task is commonly done by GPT which integrates attention method. Text summarization task is usually done with transformer model to compress a long document into a brief summary. This task is known to be one of the most difficult NLP tasks because it requires a variety of abilities, such as understanding long passages and generating consistent text that covers the core topic of the entire document. Specifically, transformer models can be prone to generating factual errors, lacking context, and oversimplifying words. [2] For that reason pointer networks which are better at extracting specific information from medical tasks,

In this paper, we propose a novel approach to biomedical text summarization that addresses the limitations of transformer models. Our approach is based on the GPT model, which is a transformer-based language model that has been shown to be effective for a variety of NLP tasks. However, we modify the GPT model by replacing the attention mechanism with a pointer network. Pointer networks are a type of neural network that are capable of explicitly attending to specific words or phrases in a document. This allows pointer networks to preserve the core values of the original text during the summarization process.

We evaluate the effectiveness of our proposed approach using the ROUGE score. ROUGE is a widely used metric for evaluating the quality of text summaries. [4] Our results demonstrate that our proposed approach outperforms the original GPT model on the ROUGE metric. These findings suggest that pointer networks can be a valuable addition to EMR systems and can provide clinicians with more accurate and informative summaries of patient medical records.

## 2. Related Work

### 2.1 Neural abstractive summarization

Abstractive sentence summarization based on a local attention-based model.[8] The model is able to generate concise summaries that convey the most important information in the input sentence, while being able to learn complex relationships between the input and output texts. The model is also structurally simple and can be easily trained end-to-end. It scales well to large amounts of training data and achieves state-of-the-art results on the DUC-2004 shared task. It demonstrates the potential of local attention-based models for abstractive text summarization, and it can be used as a baseline for future research in this area.

It also indicates some of the limitations of their model and suggest some directions for future work. One limitation is that the model can sometimes generate grammatically incorrect summaries. Another limitation is that the model is currently only able to generate sentence-level summaries. To address these limitations, the authors suggest exploring data-driven methods for improving the grammaticality of the summaries and developing methods for generating paragraph-level summaries. Both of these tasks pose additional challenges in terms of efficient alignment and consistency in generation.

### 2.2 Seq2seq

Sequence-to-sequence (seq2seq) models have been widely used for text summarization.[6] These models typically consist of an encoder and a decoder. The encoder is responsible for encoding the input text into a latent representation, while the decoder is responsible for generating the summary based on the latent representation.

A variety of seq2seq models have been proposed for text summarization, such as attention-based models [9], pointer-generator networks, and transform er models. These models have achieved promising results on a variety of summarization datasets.

However, seq2seq models can be prone to generating factual errors and lacking context. This is because seq2seq models are typically trained on a large corpus of text data. As a result, they can learn to generate summaries that are statistically likely, but not necessarily factually accurate.

### 2.3 Generative Pre-training Transformer (GPT) in summarization

Generative Pre-training Transformers (GPTs) are effective for text summarization because they are trained on a massive dataset of text and code, enabling them to learn the statistical relationships between words and phrases. This knowledge is used to generate fluent and informative summaries.

GPTs are advantageous for summarization because they can generate more abstractive summaries than other methods. Abstractive summarization involves identifying the main ideas in a text and rephrasing them in a concise and informative way. GPTs can do this because they are able to learn the underlying meaning of text.

In addition to being able to generate abstractive summaries, GPTs are also able to generate summaries that are more fluent and readable than those generated by other methods. This is because GPTs are trained on a massive dataset of text, which allows them to learn the natural patterns of human language.

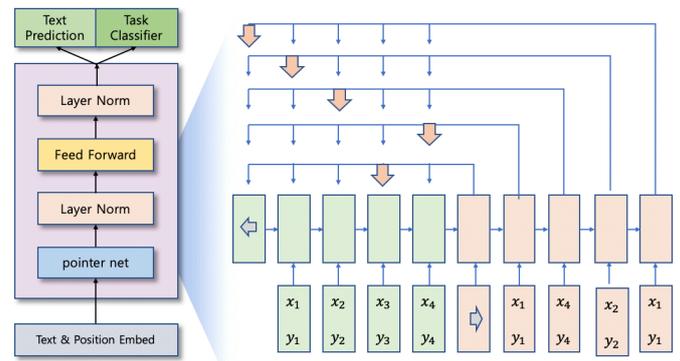

### 3. Method

Our proposed approach is based on the GPT model, which is a transformer-based language model that has been shown to be effective for a variety of natural language processing (NLP) tasks. However, we modify the GPT model by replacing the attention mechanism with a pointer network. The pointer network is responsible for selecting the words that are most relevant to the summarization task. The selected words are then used to generate the summary.

The pointer network is implemented as a neural network that takes as input the hidden states of the encoder and decoder. The pointer network then calculates a probability distribution over the words in the input text. This probability distribution is used to select the words that are most likely to be relevant to the summary. The selected words are then used to generate the summary.

In order to evaluate the model, we used medical description "Case Study: 33-Year-Old Female Presents with Chronic SOB and Cough"[7]. We modified the attention mechanism of the open-source GPT-2 model to a pointer network. We evaluated the performance of the modified model on a dataset of medical cases. The pointer GPT model outperformed the original GPT-2 model on both ROUGE(Recall-Oriented Understudy for Gisting Evaluation) scores 1 and 2, which are metrics for evaluating the quality of text summaries.

### Table 1. ROUDGE Score

| Algorithm | Evaluation metric | Precision | Recall | F measure |
|---|---|---|---|---|
| GPT2 | Rouge-1 | 0.2857 | 0.3529 | 0.3157 |
|  | Rouge-2 | 0.1 | 0.125 | 0.1111 |
| **PointerGPT** | Rouge-1 | 1.0 | 0.4705 | 0.6399 |
|  | Rouge-2 | 0.8571 | 0.375 | 0.5217 |

## 4. Conclusion

In this study, we used pointer GPT to overcome the lack of context and oversimplification, which are important limitations of attention and are essential for understanding patient information. For that, we were able to maintain the original intent of the data even after the summary processing.

Our research has the potential to usher in a new paradigm in EMR systems. By providing clinicians with more accurate and informative summaries of patient medical records, our approach can help to improve the quality of care and reduce the risk of medical errors.

## References

[1] Afzal, Muhammad, et al. "Clinical context-aware biomedical text summarization using deep neural network: model development and validation." Journal of medical Internet research 22.10 (2020): e19810.
[2] Vaswani, Ashish, et al. "Attention is all you need." Advances in neural information processing systems 30 (2017).
[3] Vinyals, Oriol, Meire Fortunato, and Navdeep Jaitly. "Pointer networks." Advances in neural information processing systems 28 (2015).
[4] Khandelwal, Urvashi, et al. "Sample efficient text summarization using a single pre-trained transformer." arXiv preprint arXiv:1905.08836 (2019).
[5] See, Abigail, Peter J. Liu, and Christopher D. Manning. "Get to the point: Summarization with pointer-generator networks." arXiv preprint arXiv:1704.04368 (2017).
[6] Sriram, Anuroop, et al. "Cold fusion: Training seq2seq models together with language models." arXiv preprint arXiv:1708.06426 (2017).
[7] case study Sharma S, Hashmi MF, Rawat D. Case Study: 33-Year-Old Female Presents with Chronic SOB and Cough. [Updated 2023 Feb 20]. In: StatPearls [Internet]. Treasure Island (FL): StatPearls Publishing; 2023 Jan-. Available from: https://www.ncbi.nlm.nih.gov/books/NBK500024/
[8] Sumit Chopra, Michael Auli, and Alexander M Rush. 2016. Abstractive sentence summarization with attentive recurrent neural networks. In North American Chapter of the Association for Computational Linguistics.
[9] Chorowski, Jan K., et al. "Attention-based models for speech recognition." Advances in neural information processing systems 28 (2015)